\documentclass[11pt]{article}
\usepackage{acl2015}
\usepackage{times}
\usepackage[hyphens]{url}
\usepackage{latexsym}
\usepackage[T1]{fontenc}
\usepackage{multirow}
\usepackage{graphicx}
\usepackage{booktabs}
\usepackage{changepage}
\usepackage{caption}
\usepackage{subcaption}
\usepackage{textcomp}
\usepackage{xcolor}
\usepackage{listings}
\title{Can GPT models be Financial Analysts? 

An Evaluation of \texttt{ChatGPT} and \texttt{GPT-4} on mock CFA Exams}

\author{\normalfont \normalsize
Ethan Callanan\textsuperscript{1,†}, Amarachi Mbakwe\textsuperscript{2,†,‡}, Antony Papadimitriou\textsuperscript{3,†}, Yulong Pei\textsuperscript{3,†}, Mathieu Sibue\textsuperscript{3,†}, \\
\normalsize Xiaodan Zhu\textsuperscript{1}, Zhiqiang Ma\textsuperscript{3}, Xiaomo Liu\textsuperscript{3}, and Sameena Shah\textsuperscript{3}\\\\
\normalsize\textsuperscript{1}Queen’s University\\
\normalsize\textsuperscript{2}Virginia Tech\\
\normalsize\textsuperscript{3}J.P. Morgan AI Research\\
\small{\tt \textsuperscript{1}\{e.callanan,xiaodan.zhu\}@queensu.ca, \textsuperscript{2}bmamarachi@vt.edu, \textsuperscript{3}\{first.last\}@jpmchase.com}
}

\date{}

\begin{document}
\maketitle
\begin{abstract}
\fontsize{10}{12}\selectfont{
Large Language Models (LLMs) have demonstrated remarkable performance on a wide range of Natural Language Processing (NLP) tasks, often matching or even beating state-of-the-art task-specific models. This study aims at assessing the financial reasoning capabilities of LLMs. We leverage mock exam questions of the Chartered Financial Analyst (CFA) Program to conduct a comprehensive evaluation of \texttt{ChatGPT}\footnote{https://platform.openai.com/docs/models/gpt-3-5} and \texttt{GPT-4}\footnote{https://platform.openai.com/docs/models/GPT-4\\ \indent  \hspace{1ex}\textsuperscript{†}Equal contribution\\ \indent \hspace{1ex}\textsuperscript{‡}Work done while interning at J.P. Morgan AI Research} in financial analysis, considering Zero-Shot (ZS), Chain-of-Thought (CoT), and Few-Shot (FS) scenarios. We present an in-depth analysis of the models’ performance and limitations, and estimate whether they would have a chance at passing the CFA exams. Finally, we outline insights into potential strategies and improvements to enhance the applicability of LLMs in finance. In this perspective, we hope this work paves the way for future studies to continue enhancing LLMs for financial reasoning through rigorous evaluation.}
\end{abstract}

\section{Introduction}

NLP has undergone a profound transformation driven by the emergence of LLMs. Models such as LLaMA \cite{touvron2023llama}, PaLM \cite{chowdhery2022palm}, \texttt{ChatGPT} and \texttt{GPT-4} from OpenAI, have garnered significant attention from both the research community and the public thanks to their impressive dialog style. From text summarization \cite{barros2023leveraging,zhang2023extractive} and code generation \cite{le2022coderl,yang2023exploitgen} to question answering \cite{wang2022numerical,khashabi2020unifiedqa}, named entity recognition \cite{li2022unified}, and beyond, LLMs have showcased remarkable performance across diverse tasks. 
% TODO: fix this table throwing errors
\begin{table}[h]
\begin{tabular}{ lc|cc } 
\toprule
\textbf{Model} & \textbf{Setting} & \textbf{Level I} & \textbf{Level II} \\
\hline
\multirow{6}{4em}{\textbf{\texttt{ChatGPT}}} & ZS & $58.8\pm 0.2$ & $46.6\pm 0.6$ \\ 
& CoT & $58.0\pm 0.2$ & $47.2\pm 0.3$ \\ 
& 2S & $\textbf{63.0}\pm 0.2$ & $46.6\pm 0.1$ \\ 
& 4S & $62.3\pm 0.2$ & $45.7\pm 0.2$ \\ 
& 6S & $62.2\pm 0.2$ & $47.0\pm 0.3$ \\ 
& 10S & $62.4\pm 0.2$ & $\textbf{47.6}\pm 0.4$ \\ 
\midrule
\multirow{6}{4em}{\textbf{\texttt{GPT-4}}} & ZS & $73.2\pm 0.2$ & $57.4\pm 1.5$ \\ 
& CoT & $74.0\pm 0.2$ & $\textbf{61.4}\pm 0.9$ \\
& 2S & $73.9\pm 0.1$ & $60.2\pm 0.9$ \\ 
& 4S & $73.8\pm 0.2$ & $60.5\pm 0.7$ \\ 
& 6S & $74.5\pm 0.2$ & - \\ 
& 10S & $\textbf{74.6}\pm 0.2$ & - \\ 
\bottomrule
\end{tabular}
\caption{\label{level1overall} Overall Performance of \texttt{ChatGPT} and \texttt{GPT-4} on Level I and Level II Exams (Accuracy) in Zero-Shot (ZS), Chain-of-Thought (CoT), and Few-Shot (FS) settings.}
\end{table}

In finance, NLP has played a pivotal role in enhancing various services, such as customer relations, stock sentiment analysis, financial question answering \cite{wang2022numerical}, document understanding \cite{kim2022ocr}, and report summarization \cite{abdaljalil2021exploration}. Despite these advancements, applying NLP in finance poses unique challenges, such as the distinct nature of financial tasks, linguistic structures, and specialized terminology. As a result, the performance of general NLP models often falls short when applied to finance-related tasks -- the specific challenges of financial reasoning problems warrant further investigation.

\begin{figure*}[htp]
\begin{subfigure}{\textwidth}%[h]
    \fontsize{9}{11}\selectfont{
    \textbf{Which of the following is most likely an assumption of technical analysis?}\\
    \textit{\textcolor{blue}{A. Security markets are efficient\\
    B. Market trends reflect irrational human behavior\\
    C. Equity markets react quickly to inflection points in the broad economy}}
    }
    \caption{Level I sample question}
\end{subfigure}%

\begin{subfigure}{\textwidth}%[h]
    \fontsize{9}{11}\selectfont{
    \textit{\textcolor{orange}{\\Paris Rousseau, a wealth manager at a US-based investment management firm, is meeting with a new client. The client has asked Rousseau to make recommendations regarding his portfolio’s exposure to liquid alternative investments [...] \\ {[Table Evidence]}
    }}\\
    \textbf{The AFFO per share for Autier REIT over the last 12-months is closest to: }\\
    \textit{\textcolor{blue}{A. \$6.80. \\
    B. \$7.16. \\
    C. \$8.43. }}
    }
    \caption{Level II sample question}
\end{subfigure}
\caption{CFA example questions (source: CFA Institute); the question appears in bold, the multiple choices in blue and italic, and the vignette/case description in orange and italic}
\label{fig:cfa-examples}
\end{figure*}

In this paper, we rigorously assess the capabilities of LLMs in real-world financial reasoning problems by conducting an evaluation on mock exam questions of the prestigious Chartered Financial Analyst (CFA) Program\footnote{\url{https://www.cfainstitute.org/en/programs/cfa/exam} \\ \indent \hspace{1ex}Code available at https://github.com/e-cal/gpt-cfa}. The CFA exams are known for their meticulous yet practical assessment of financial expertise, making their resolution an ideal use case to gauge the capabilities of LLMs in handling complex financial reasoning scenarios. Our work focuses on two closed-source, non-domain specific LLMs, \texttt{ChatGPT} and \texttt{GPT-4}, using various popular prompting techniques. Our contributions are as follows:

\begin{itemize}
\item[1] We conduct the first comprehensive evaluation of \texttt{ChatGPT} and \texttt{GPT-4} in financial reasoning problems using CFA mock exam questions, considering ZS, CoT, and FS scenarios.
\item[2] We present an in-depth analysis of the models' performance and limitations in solving these financial reasoning problems, and estimate how they would fare in the Level I and Level II CFA exams.
\item[3] We outline insights into potential strategies and improvements to enhance the applicability of LLMs in finance, opening new avenues for research and development.
\end{itemize}

\section{Related Work} 

\subsection{LLMs and Finance} 
LLMs are Transformer-based generative models \cite{vaswani2017attention} trained on massive datasets that cover a broad range of topics and domains. Previous work has demonstrated the ability of LLMs to generalize surprisingly well to unseen downstream tasks, with little to no additional training data \cite{brown2020language,wei2022chain}. This raises an interesting question on the competitiveness of LLMs with supervised state-of-the-art models on specialized domains, such as finance. Indeed, the characteristics of most financial tasks — which rely on very specific concepts and mathematical formula, frequently leverage diagrams and tables, often need multistep reasoning with calculations — make finance a challenging domain of application for LLMs. Several paths have been proposed to incorporate or emphasize domain-specific knowledge in LLMs: continued pre-training \cite{araci2019finbert,wu2023bloomberggpt} and supervised fine-tuning on new data \cite{mosbach2023few,yang2023fingpt}, retrieval augmented generation using a vector database of external knowledge \cite{lewis2020retrieval}, etc. However, before considering such enhancements, only few papers have proceeded to extensively benchmark the out-of-the-box capabilities of newer instruction-tuned LLMs \textit{in finance} \cite{li2023chatgpt}.

\begin{table*}[h]
\small
\begin{adjustwidth}{-2em}{}
\centering
\begin{tabular}{l|cccccccccc}
\toprule
 & \multicolumn{3}{c}{\textbf{Level I}} & & \multicolumn{3}{c}{\textbf{Level II}} \\
 \cmidrule{2-4} \cmidrule{6-8}
\textbf{Topic} & \textbf{Calculations} & \textbf{\#Tables} & \textbf{Len(Prompt)} & & \textbf{Calculations} & \textbf{\#Tables} & \textbf{Len(Prompt)} \\
\midrule
Ethics & 0.7\% & 0.01 & 125 & & 0.0\% & 0.00 & 1013 \\
Derivatives & 20.7\% & 0.00 & 65 & & 75.0\% & 2.00 & 816 \\
Alternative Investments & 36.4\% & 0.06 & 85 & & 66.7\% & 2.00 & 840 \\
Portfolio Management & 38.3\% & 0.18 & 110 & & 56.3\% & 2.13 & 1077 \\
Fixed Income & 43.0\% & 0.06 & 87 & & 50.0\% & 1.45 & 779 \\
Economics & 50.6\% & 0.25 & 121 & & 66.7\% & 2.00 & 1115 \\
Equity & 52.5\% & 0.19 & 112 & & 45.8\% & 1.00 & 1053 \\
Corporate Issuers & 59.3\% & 0.28 & 120 & & 44.4\% & 1.67 & 930 \\
Quantitative Methods & 70.5\% & 0.26 & 131 & & 27.8\% & 0.00 & 1256 \\
Financial Reporting & 57.7\% & 0.35 & 151 & & 53.6\% & 2.79 & 1383 \\
\midrule
\textbf{Overall} & 42.4\% & 0.17 & 116 & & 45.5\% & 1.47 & 1058 \\
\bottomrule
\end{tabular}
\end{adjustwidth}
\caption{\label{q-characteristics}Question characteristics by topic; percentage of questions requiring calculation, average number of table evidence per question, and average prompt length (estimated using the tiktoken Python package)}
\end{table*}

\subsection{Evaluation of LLMs on Human Exams and other Benchmarks}
Several previous studies have evaluated the performance of LLMs on different standard exams. Tests considered include the United States medical licensing exam \cite{kung2023performance}, free-response clinical reasoning exams \cite{strong2023performance}, college-level scientific exams \cite{wang2023scibench}, the Bar exam \cite{katz2023gpt}, the driver’s license knowledge test \cite{rahimi2023exploring}, and more. The crucial contribution that these works bring to the scientific community and the industry is an in-depth analysis of the strengths and weaknesses of LLMs in realistic domain-specific settings. Through their conclusions, such investigations guide subsequent research and practical use case resolutions in industry. 

For example, \cite{wang2023scibench} evaluated \texttt{ChatGPT} and \texttt{GPT-4} on a collection of Physics, Chemistry, and Math problems, and then concluded that current LLMs do not deliver satisfactory performance in complex scientific reasoning yet to be reliably leveraged in practice. In contrast, \cite{bang2023multitask} found that \texttt{ChatGPT} outperformed fine-tuned task-specific models on four different NLP tasks, thus suggesting \texttt{ChatGPT} could be directly applied to solve industry use cases involving these tasks. 

Our paper aims at following the footsteps of \cite{li2023chatgpt} and delves further into the assessment of the inner financial reasoning abilities of \texttt{ChatGPT} and \texttt{GPT-4} to help future industry applications.

\section{Dataset}

The CFA Program is a three-part exam that tests the fundamentals of investment tools, valuing assets, portfolio management, and wealth planning. It is typically completed by those who want to work in the financial industry with backgrounds in finance, accounting, economics, or business. Successfully completing the CFA Program reflects a strong grasp of fundamental financial knowledge, and charterholders are then qualified for roles related to investment management, risk management, asset management, and more.

As mentioned above, the CFA exam is composed of three levels, each with a specific format. Irrespective of the level, each problem from the CFA exam is affiliated to one of ten distinct finance topics: Ethics, Quantitative Methods, Economics, Financial Statement Analysis, Corporate Issuers, Portfolio Management, Equity Investments, Fixed Income, Derivatives, and Alternative Investments. Level I features a total of 180 independent Multiple Choice Questions (MCQs). Level II consists of 22 item sets comprised of vignettes (i.e., case descriptions with evidence) and 88 accompanying MCQs. Finally, Level III comprises a mix of vignette-supported essay questions and vignette-supported multiple choice questions. 

Two main challenges arise when trying to benchmark any model on the CFA exam. Firstly, the CFA Institute refrains from publicly releasing past exams taken by registered candidates, making the collection of official questions and answers directly from any CFA exam impossible. Secondly, a significant fraction of the level III item sets expects plain text responses, which then require the costly intervention of human experts for grading. To circumvent these difficulties, we decide to rely on mock CFA exams and choose to solely focus on levels I and II, leaving Level III to future work. We collected a total of five Level I mock exams and two Level II mock exams. We share in Figure \ref{fig:cfa-examples} example MCQs published by the CFA Institute for Level I and Level II. We ensure each topic is represented in similar proportions to the original CFA sections (Figure \ref{fig:l1-topic-dist} and Figure \ref{fig:l2-topic-dist} in the Appendix). %Figures ZZZ and WWW illustrate the topic distribution of our Level I and Level II internal datasets.% 
Table \ref{q-characteristics} summarizes important statistics about Level I and Level II problems.

\section{Experiments}

\subsection{Setup}

\begin{table*}[h]
\small
\begin{center}
\begin{tabular}{l|ccccccc}
\toprule
 & \multicolumn{3}{c}{\textbf{\texttt{ChatGPT}}} & & \multicolumn{3}{c}{\textbf{\texttt{GPT-4}}} \\
 \cmidrule{2-4} \cmidrule{6-8}
\textbf{Category} & \textbf{ZS} & \textbf{CoT} & \textbf{2S} & & \textbf{ZS} & \textbf{CoT} & \textbf{10S} \\
 \midrule
Ethics & $59.2 \pm 0.1$ & $59.2 \pm 1.4$ & $\textbf{64.6}\pm 0.9$ & & $80.3 \pm 0.7$ & $78.9\pm 0.4$ & $\textbf{82.4} \pm 0.5$\\
Quantitative Methods & $53.9 \pm 0.2$ & $50.0\pm 0.8$ & $\textbf{59.7} \pm 1.0$ & & $\textbf{78.0}\pm 0.7$ & $76.0\pm 1.1$ & $76.0 \pm 0.8$\\
Economics & $\textbf{68.0}\pm 1.1$ & $63.7\pm 2.5$ & $\textbf{68.0} \pm 3.9$ & & $74.1 \pm 1.9$ & $73.6\pm 1.2$ & $\textbf{76.2} \pm 0.6$\\
Financial Reporting & $54.0 \pm 1.2$ & $53.4\pm 0.6$ & $\textbf{60.1} \pm 0.7$ & & $68.2\pm 1.0$ & $\textbf{70.8}\pm 1.3$ & $70.0 \pm 0.7$\\
Corporate Issuers & $71.4 \pm 5.2$ & $69.8\pm 4.8$ & $\textbf{74.2} \pm 4.1$ & & $74.4\pm 4.1$ & $74.6 \pm 6.2$ & $\textbf{75.3} \pm 4.0$\\
Equity Investments & $59.4\pm 0.1$ & $60.9 \pm 0.7$ & $\textbf{62.5} \pm 1.0$ & & $\textbf{80.3}\pm 0.7$ & $70.5\pm 0.9$ & $68.8 \pm 0.8$\\
Fixed Income & $55.6\pm 1.4$ & $60.2 \pm 0.5$ & $\textbf{63.6} \pm 0.5$ & & $\textbf{74.9}\pm 2.6$ & $60.2\pm 0.5$ & $73.6 \pm 0.8$\\
Derivatives & $61.1\pm 4.1$ & $68.5 \pm 2.1$ & $\textbf{73.0} \pm 1.5$ & & $90.5\pm 0.8$ & $93.8 \pm 0.7$ & $\textbf{96.0} \pm 0.5$\\
Alternative Investments & $60.7 \pm 2.4$ & $60.7 \pm 1.9$ & $\textbf{62.9} \pm 1.1$ & & $75.9\pm 1.1$ & $\textbf{77.1}\pm 1.0$ & $72.1 \pm 1.3$\\
Portfolio Management & $58.3 \pm 2.8$ & $48.3\pm 3.6$ & $\textbf{61.7} \pm 2.4$ & & $63.7\pm 0.6$ & $71.7 \pm 0.9$ & $\textbf{79.6} \pm 1.4$\\\midrule
\bf Overall & $58.8 \pm 0.2$ & $58.0\pm 0.2$ & $\textbf{63.0} \pm 0.2$ & & $73.2\pm 0.2$ & $74.0 \pm 0.9$ &$\textbf{74.6} \pm 0.2$\\
\bottomrule
\end{tabular}
\end{center}
\caption{\label{l1-results} \texttt{ChatGPT} and \texttt{GPT-4} accuracy on Level I Exams}
\end{table*}

\begin{table*}[h]
\small
\begin{center}
\begin{tabular}{l|ccccccc}
\toprule
 & \multicolumn{3}{c}{\textbf{\texttt{ChatGPT}}} & & \multicolumn{3}{c}{\textbf{\texttt{GPT-4}}} \\
 \cmidrule{2-4} \cmidrule{6-8}
\textbf{Category} & \textbf{ZS} & \textbf{CoT} & \textbf{10S} & & \textbf{ZS} & \textbf{CoT} & \textbf{4S}\\
 \midrule
Ethics & $31.3\pm 7.6$  & $\textbf{37.5}\pm 9.5$ & $21.9\pm 4.6$ & & $43.8\pm 1.6$ & $56.3\pm 1.2$ & $\textbf{59.4} \pm 1.5$\\
Quantitative Methods & $44.4\pm 12.0$ & $\textbf{55.6}\pm 6.5$ & $54.2 \pm 9.3$ & & $66.7 \pm 1.1$ & $66.7 \pm 7.4$ & $\textbf{72.2} \pm 4.3$\\
Economics & $\textbf{66.7}\pm 0.0$ & $58.3\pm 1.4$ & $62.5 \pm 1.9$ & & $41.7\pm 1.4$ & $\textbf{58.3}\pm 6.3$ & $50.0 \pm 6.9$\\
Financial Reporting & $39.6 \pm 3.4$ & $31.3\pm 2.0$ & $\textbf{44.8} \pm 2.5$ & & $54.2\pm 3.9$ & $\textbf{66.7}\pm 4.2$ & $63.5 \pm 3.3$\\
Corporate Issuers & $\textbf{55.6}\pm 3.7$ & $50.0\pm 2.8$ & $50.0 \pm 1.9$ & & $77.8\pm 0.9$ & $77.8\pm 0.6$ & $\textbf{80.6} \pm 1.3$\\
Equity Investments & $60.4 \pm 1.6$ & $60.4 \pm 9.9$ & $\textbf{60.9} \pm 7.0$ & & $\textbf{65.0}\pm 5.7$ & $58.8\pm 7.3$ & $62.5 \pm 4.7$\\
Fixed Income & $\textbf{38.9}\pm 0.9$ & $27.8\pm 6.5$ & $34.4 \pm 1.9$ & & $60.0\pm 5.8$ & $\textbf{62.2}\pm 0.8$ & $53.9 \pm 1.9$\\
Derivatives & $50.0\pm 5.6$ & $\textbf{58.3}\pm 12.5$ & $47.9 \pm 3.1$ & & $\textbf{66.7}\pm 5.6$ & $58.3\pm 0.7$ & $50.0 \pm 4.2$\\
Alternative Investments & $33.3 \pm 0.0$ & $33.3 \pm 0.0$ & $\textbf{58.3} \pm 0.7$ & & $66.7 \pm 0.0$ & $50.0\pm 0.0$ & $\textbf{75.0} \pm 0.7$\\
Portfolio Management & $47.2\pm 0.9$ & $\textbf{66.7}\pm 8.3$ & $59.7 \pm 9.5$ & & $36.1\pm 1.6$ & $55.6 \pm 0.6$ & $\textbf{56.9} \pm 4.3$\\
\midrule
\bf Overall & $46.6\pm 0.6$ & $47.2 \pm 0.3$ & $\textbf{47.6} \pm 0.4$ & & $57.4\pm 1.5$ & $\textbf{61.4}\pm 0.9$ & $60.5 \pm 0.7$\\
\bottomrule
\end{tabular}
\end{center}
\caption{\label{l2-results} \texttt{ChatGPT} and \texttt{GPT-4} accuracy on Level II Exams}
\end{table*}

This section outlines the methodology employed to assess the financial reasoning abilities of \texttt{ChatGPT} and \texttt{GPT-4} using mock CFA exams. Our study examined various prompting paradigms.\\
\textbf{ZS prompting:} We gauged the models' inherent reasoning abilities without providing any correct examples in the input. \\
\textbf{FS prompting:} We furnished the models with prior examples of expected behavior to facilitate the acquisition of new knowledge that could aid in solving the questions. We tested two different strategies to select FS examples: (a) randomly sampling from the entire set of questions within the exam level (2S, 4S and 6S), and (b) sampling one question from each topic in the exam level (10S). This last approach aims at enabling the models to discern the distinct attributes of each topic within every exam level. Due to the limited context window of \texttt{GPT-4} and the length of the Level II item-sets (case description and question), 6S and 10S prompting were not evaluated for \texttt{GPT-4} on the Level II mock exams.\\
\textbf{CoT prompting:} For each exam level, we also evaluated the models by prompting them to think through the input problem step-by-step and show their work for calculations (also known as ZS CoT) \cite{wei2022chain}. This has the added benefit of allowing us to analyze the models' "problem-solving process" and thus determine where and why it might have gone wrong.\\
\textbf{Implementation Details:} We conducted the experiments using the OpenAI ChatCompletion API (\texttt{gpt-3.5-turbo} and \texttt{gpt-4} models) with functions and set the temperature parameter to zero, thereby eliminating randomness in the models' generations. The prompt templates we crafted for each level and for each prompting setting can be found in the Appendix. We employed a memorization test as in \cite{kiciman2023causal} to confirm that the models had not memorized the mock exams as part of their training data. \\
\textbf{Metrics:} To measure the performance of LLMs on the mock exam MCQs, we compared their predictions against the established solution set of each of the CFA mock exams collected. Accuracy served as our sole evaluation metric throughout this study.

\subsection{Results Overview}

\textbf{LLMs struggle more on Level II than on Level I:} We notice that, no matter the prompting paradigm employed, both \texttt{ChatGPT} and \texttt{GPT-4} encounter more difficulties correctly answering the item-sets from Level II than the independent questions from Level I (Table \ref{l1-results}, Table \ref{l2-results}). While there is no general consensus as to which level is usually considered harder for exam takers, we suggest that three factors might have negatively affected the performance of LLMs in Level II based on our analysis. 

Firstly, the case description attached to each item-set from Level II increases the length of the input prompt and dilutes the useful information it contains. Indeed, we observe that Level II prompts are on average ${\raise.17ex\hbox{$\scriptstyle\sim$}}10\times$ longer than Level I prompts; confronting Table \ref{q-characteristics}, Table \ref{l1-results}, Table \ref{l2-results} shows that topics associated with poor performance usually present longer contexts both in Level I and Level II. In addition, the detailed case descriptions from Level II depict realistic day-to-day situations that contrast with the more general questions from Level I: LLMs thus need to abstract from case-specific details in Level II questions so as to identify the underlying finance concepts involved. 

Secondly, as Level II questions are grouped into item-sets, each item tends to go more in-depth about a specific finance topic than the questions that compose Level I, thus leading to more specialized and intricate problems. 

Lastly, the Level II section features a slightly higher proportion of questions requiring calculations and a much higher proportion of questions containing table evidence, in comparison to Level I (Table \ref{q-characteristics}). Given the known limitations of LLMs for out-of-the-box numerical and table reasoning \cite{frieder2023mathematical,chen2022convfinqa}, this could also explain the lower accuracy observed in Level II across the board.\\

\noindent{\textbf{\texttt{GPT-4} outperforms \texttt{ChatGPT} in almost all experiments, but certain finance topics remain challenging for both models:} As shown in Table \ref{l1-results} and Table \ref{l2-results}, \texttt{GPT-4} consistently beats \texttt{ChatGPT} in all topics in Level I and most topics in Level II, irrespective of the learning paradigm. }

In Level I, we see that both LLMs perform best in the Derivatives, Alternative Investments, Corporate Issuers, Equity Investments, and Ethics topics. For Derivatives and Ethics, this observation can be explained by the low amount of calculations and table understanding required to answer correctly (Table \ref{q-characteristics}). The explicit mention of popular finance notions in the questions of Derivatives and Ethics (e.g., options, arbitrage, etc.) further reduces their difficulty too. Similarly, in Alternative Investments, Corporate Issuers, and Equity Investments, problems often directly refer to well-known finance concepts that might have been encountered by \texttt{ChatGPT} and \texttt{GPT-4} during pretraining or instruction-tuning -- thus facilitating their resolution despite having more calculations involved. However, both models show relatively poor performance in the Financial Reporting and Portfolio Management topics in Level I, with \texttt{ChatGPT} also struggling a lot more on highly computational topics such as Quantitative Methods. Indeed, Portfolio Management and Financial Reporting problems are more case-based, applied, computational, and CFA-specific than the ones from the aforementioned topics, which might have negatively affected performance. They also tend to include more table evidence and complex details to leverage (Table \ref{q-characteristics}). 

In Level II, we observe that both \texttt{ChatGPT} and \texttt{GPT-4} still perform relatively strongly on Derivatives, Corporate Issuers, and Equity Investments, yet still relatively poorly on Financial Reporting. However, the results are now more nuanced: \texttt{ChatGPT} struggles on Alternative Investments and Fixed Income compared to \texttt{GPT-4}, while \texttt{ChatGPT} outperforms \texttt{GPT-4} in Portfolio Management and Economics. Interestingly enough, both models now demonstrate low answer accuracy in the Ethics item-sets of Level II. This could originate from the more in-depth, situational, and detailed character of the problems from Level II in comparison to Level I. \\

\noindent{\textbf{CoT prompting yields limited improvements over ZS:} Although CoT performs better than ZS in almost all cases and better than FS in Level II for \texttt{GPT-4}, we note that the use of CoT did not help LLMs as much as we initially expected (Table \ref{level1overall}, Table \ref{l1-results}, Table \ref{l2-results}). In Level I, CoT prompting hardly benefits \texttt{GPT-4} (bringing in just a 1\% relative increase) and actually deteriorates the performance of \texttt{ChatGPT}. In Level II, CoT prompting yields a decent 7\% relative improvement over ZS prompting for \texttt{GPT-4}, but a disappointing 1\% for \texttt{ChatGPT}. Section 5.1 further investigates the reasons explaining such observations. In Level I, we see that CoT negatively affected both LLMs particularly in Quantitative Methods, which could be due to hallucinations in mathematical formula and calculations. In Level II, we notice that CoT benefited both LLMs in the Ethics and Portfolio Management topics, where explicit step-by-step reasoning over long and intricate evidence is usually helpful. In both levels, we also noted that CoT prompting sometimes led to inconsistent performance across questions from the same topic, as manifested by the high standard deviations reported in Table \ref{l1-results} and Table \ref{l2-results}. }

However, despite the aforementioned observations, it is hard to clearly identify  more topics that systematically benefit or suffer from the use of CoT for both models across levels. For instance, in Financial Reporting problems from Level II, \texttt{GPT-4} saw its accuracy improve by 23\% with CoT relative to ZS, while \texttt{ChatGPT} saw its performance decrease by 21\% (Table \ref{l2-results}). \\

\noindent{\textbf{A few in-context examplars help more than CoT:}
Compared to ZS and CoT prompting, FS prompting offers significant performance improvements for \texttt{ChatGPT} on the Level I mock exams (Table \ref{level1overall}). 2S prompting yielded the best performance across all categories and overall in Level I for \texttt{ChatGPT}. Across mock exams in Level II, the dominance is not as significant, but FS prompting still manages to achieve the best overall score for both models, with the exception of Level II for \texttt{GPT-4} (Table \ref{l1-results}, Table \ref{l2-results}). Interestingly, for Level II, the best FS prompting type was 10S prompting for \texttt{ChatGPT}, which suggests  more complex exams benefited from a more holistic FS approach across multiple topics. The overall trend shown in the results is that FS prompting seems to offer better assistance to  less complex models (\texttt{ChatGPT}) when being tested on seemingly simpler exams (Level I).

It is likely that FS yields better performance improvement than CoT because it shows actual correct answers to different types of mock questions. It also enables the models to understand how to best use the table evidence or other information contained in a question (if any). The comparatively lower performance improvement brought by FS observed in Level II mock exams may be due to the more complex nature of the questions and the fact they include case studies; it may be a scenario where simply prompting the models with the correct answers is not sufficient. Level II may thus benefit from a combination of FS and CoT prompting with clear explanations as to how the information in the case study was leveraged to arrive at the correct answer.}

\section{Detailed Analysis}

\subsection{Underperformance of CoT on Level I}

It was surprising to see that CoT only marginally improved the models' performance on each test, and was actually slightly detrimental to the performance of \texttt{ChatGPT} on the Level I exams. To inspect the nature of the errors made by the models when using CoT prompting, we looked over each instance where no-CoT was correct while CoT was incorrect, and categorized the error as one of: Knowledge, Reasoning, Calculation, or  Inconsistency. 

Knowledge errors are those where the model lacks critical knowledge required to answer the question. This includes an incorrect understanding of some concept, not knowing the relationship between concepts, or using an incorrect formula to answer a question requiring calculation. Reasoning errors are when the model had all the correct knowledge, but either over-reasoned in its response, or hallucinated some additional requirements or information in the question that was not actually present. Calculation errors are errors pertaining to some incorrect calculation (using a correct formula), or failing to accurately compare or convert results. Errors of inconsistency are when the model's thinking is entirely correct, yet it chooses the wrong answer.

\begin{table}[h]
\begin{center}
\begin{tabular}{l|cc}
\toprule
 % & \multicolumn{1}{c}{\textbf{\texttt{ChatGPT}}} & \multicolumn{1}{c}{\textbf{\texttt{GPT-4}}} \\
 % \cmidrule{2-2} \cmidrule{3-3}
\textbf{Type of Error} & \textbf{\texttt{ChatGPT}} & \textbf{\texttt{GPT-4}} \\
 \midrule
Knowledge & $55.2\%$ & $50.0\%$ \\
Reasoning & $8.6\%$ & $10.7\%$ \\
Calculation & $17.2\%$ & $28.6\%$ \\
Inconsistency & $19.0\%$ & $10.7\%$ \\
\bottomrule
\end{tabular}
\end{center}
\caption{\label{l1-error-modes} Error modes of level I questions \texttt{ChatGPT} and \texttt{GPT-4} got correct without CoT but incorrect using CoT}
\end{table}

\noindent{\textbf{\texttt{ChatGPT}:} By far the most common error mode for \texttt{ChatGPT} is knowledge based, constituting over half of all errors VS. no-CoT. This implies that, with CoT reasoning, the gaps in the LLMs internal knowledge are magnified. As the model begins to think through its answer, it states its incorrect assumptions, which it proceeds to rationalize in the context of the question thereby skewing the rest of the answer towards a wrong choice. Without using CoT reasoning, the model is able to make an "educated guess" where any incorrect knowledge has less of an opportunity of skewing the guess towards an incorrect answer. With a 1/3 chance of guessing correctly, plus any contextual hints that may lie in the question, for questions where GPT simply lacks the knowledge to reason correctly, guessing is a more accurate strategy.}

This same principal similarity explains calculation and reasoning errors, where one or a few off-track token generations then throw off the rest of the answer, resulting in an incorrect conclusion. 

The instances where the model is entirely correct but then concludes or just selects the wrong answer are more enigmatic. In about half of these cases, it seems to fail to generate a stop token upon coming to the conclusion, leading it to restate the concluding sentence with another option selected. In the other cases, there appears to be some disconnect between the thought process and the answer selection. As we were using OpenAI's functions API to retrieve structured output, our leading suspicion is that in these cases the ordering outlined in the system prompt was missed or ignored, and the answer was generated first.

% \subsubsection{\texttt{GPT-4}}

\noindent{\textbf{\texttt{GPT-4}:} There were about half as many instances of CoT making an error not made without CoT for \texttt{GPT-4}, compared to \texttt{ChatGPT}. On these questions, \texttt{GPT-4} also displays knowledge errors as the most common error mode. However, unlike \texttt{ChatGPT}, almost none of these knowledge errors were using the incorrect formula. This, along with the fact that there were less knowledge errors in total, shows that \texttt{GPT-4} has more complete internal knowledge of both financial information and especially financial formulas and calculation methods. Rather than knowledge errors, \texttt{GPT-4}'s most common error mode on questions requiring calculation are calculation errors. \texttt{ChatGPT} also frequently made these sorts of errors in conjunction with using the wrong formula, which underlines the well-known and more foundational shortcoming of language models' mathematical abilities \cite{frieder2023mathematical}.}

\texttt{GPT-4} also displayed far fewer inconsistency errors than \texttt{ChatGPT}. It appears to have a much stronger ability to connect its reasoning to the answers and to make comparisons. The one error type that \texttt{GPT-4} makes more frequently than \texttt{ChatGPT} was reasoning errors. It would seem that, along with \texttt{GPT-4}'s greater ability to reason, it has a greater chance of "talking itself" into incorrect lines of reasoning.

\begin{table}[h]
\begin{center}
\begin{tabular}{l|cc}
\toprule
 % & \multicolumn{1}{c}{\textbf{\texttt{ChatGPT}}} & \multicolumn{1}{c}{\textbf{\texttt{GPT-4}}} \\
 % \cmidrule{2-2} \cmidrule{3-3}
\textbf{Type of Error} & \textbf{\texttt{ChatGPT}} & \textbf{\texttt{GPT-4}} \\
 \midrule
Knowledge     & $70\%$ & $80\%$ \\
Reasoning     & $20\%$ & $20\%$ \\
Out of Tokens & $10\%$ & $0\%$ \\
\bottomrule
\end{tabular}
\end{center}
\caption{\label{l2-error-modes} Error modes of level II questions \texttt{ChatGPT} and \texttt{GPT-4} got correct without CoT but incorrect using CoT}
\end{table}

\subsection{CoT Benefits on Level II}

If CoT amplifies the effect of missing knowledge, and allows LLMs room to miscalculate or "talk themselves" into a wrong answer, one might question why it seemed to help much more on Level II exams. The Level II exam questions require more interpretation of the information, as one needs to figure out what is relevant from the case, and some information may be missing but is expected to be known and needed to answer the question. Using CoT helps the model to reason over the information and filter what is relevant to the question from the case.

\subsection{Can LLMs pass the CFA exam?}
\subsubsection{CFA Level I Passing Score}
The CFA Institute refrains from disclosing the minimum passing score (MPS) for its examinations, thereby giving rise to an entire industry centered around speculating on the elusive actual MPS. The MPS is uniquely established for each individual exam, guided by the standards that the CFA Institute established back in 2011.

The CFA Institute employs the `Angoff Standard Setting Method' to ascertain the pass rates for CFA exams. This involves a group of CFA Charterholders convening to collectively assess the true difficulty level of the questions and the appropriate level of ease that should accompany passing each question.

\begin{table*}[h]
\begin{center}
\begin{tabular}{l|ccccccc}
\toprule
 & \multicolumn{3}{c}{\textbf{\texttt{ChatGPT}}} & & \multicolumn{3}{c}{\textbf{\texttt{GPT-4}}} \\
 \cmidrule{2-4} \cmidrule{6-8}
\textbf{Exam} & \textbf{ZS} & \textbf{CoT} & \textbf{FS} & & \textbf{ZS} & \textbf{CoT} & \textbf{FS} \\
 \midrule
Level I & Pass & Fail & Pass & & Pass & Pass & Pass\\
Level II & Fail & Fail & Fail & & Unclear & Pass & Pass\\
\bottomrule
\end{tabular}
\end{center}
\caption{\label{pass-rates} \texttt{ChatGPT} and \texttt{GPT-4} ability to pass Level I and Level II Exams}
\end{table*}

Although the CFA Institute maintains an air of secrecy surrounding its pass/fail thresholds, certain indicators point towards a potential elevation of the MPS for CFA Level I. Drawing from feedback provided by CFA exam takers on Reddit, the average MPS stood at 65\% in December 2019, but surged to 71.1\% by February 2021. 
% On the lower end of the spectrum, a Reddit user overseeing the compilation of scores speculates that the minimum passing score has also witnessed an increase. 
In June 2019, estimations suggest that certain individuals managed to pass CFA Level I with a mere 60.8\%; by February 2021, this had escalated to 66.7\%.

Aiming for approximately 70\% in as many subjects as possible seems to be a prudent strategy for clearing CFA Level I. Put differently, attaining scores above 70\% in all topics is not a necessity for passing. Some contend that achieving as low as 65\% or even 63\% might suffice. Remarkably, one doesn't even need to exceed 51\% in every area to secure a passing grade. The pattern appears to allow for the possibility of scoring below 50\% in about three, or perhaps four, subjects. However, this would likely necessitate counterbalancing with scores exceeding 70\% in at least three subjects and falling between 51\% and 70\% in the remaining ones. Nevertheless, maintaining an average score of 70\% across subjects considerably enhances the likelihood of a positive outcome upon receiving the results.
\footnote{\url{https://www.efinancialcareers.com.au/news/finance/whats-the-minimum-score-you-can-get-on-cfa-level-i-and-still-pass}}

\subsubsection{CFA Level II Passing Score}
The estimations from the Reddit community regarding the MPS for CFA Levels II and III are even more outdated than those for Level I, yet they indicate that the two advanced exams have consistently featured lower passing thresholds. In June 2019, their approximations pegged the MPS for Level III at a mere 57.4\%, and for Level II at just 62.8\%. The subject level passing scores are ambiguous for the Level II exam, but we can attempt to apply the same logic as the Level I exam but make an assumption that threshold for each subject is 60\% instead of 70\%.\footnote{\url{https://www.efinancialcareers.com.au/news/finance/whats-the-minimum-score-you-can-get-on-cfa-level-i-and-still-pass}}

\subsubsection{Proposed pass criteria}
Given the above information our proposed pass criteria is as follows:
\begin{itemize}
  \item Level I - achieve a score of at least 60\% in each topic and an overall score of at least 70\%
  \item Level II - achieve a score of at least 50\% in each topic and an overall score of at least 60\%
\end{itemize}

Table \ref{pass-rates} shows which model implementations were able to pass the exams. The FS implementations in both settings correspond to the number of shots shown in Table \ref{l1-results} and Table \ref{l2-results}. Most of the settings showed a clear pass or fail except for \texttt{GPT-4} ZS on Level II which was a borderline decision either way. \texttt{GPT-4} in a ZS setting attains a score of >60\% in six of the topics and achieves a score of between 50\% and 60\% in one of the topics. The topic performance seems high but the overall score of 57.39\% falls slightly short of the minimum passing score proposed earlier, it is thus unclear as to whether this LLM setting would pass the CFA Level II exam.

\section{Conclusion and Discussion}
\label{sect:Conclusion and Discussion}

In this paper, we have conducted a thorough evaluation of \texttt{ChatGPT} and \texttt{GPT-4} on the CFA level I and level II exams. We observed that \texttt{GPT-4} performed better than \texttt{ChatGPT} in almost every topic of both levels when using the same prompting method. Based on estimated pass rates and average self-reported scores, we concluded that \texttt{ChatGPT} would likely not be able to pass the CFA level I and level II under all tested settings, while \texttt{GPT-4} would have a decent chance of passing the CFA Level I and Level II if prompted with FS and/or CoT.

We noted that CoT prompting provided little improvement for \texttt{ChatGPT} on both exams and \texttt{GPT-4} on the Level I exam. While CoT prompting did help the models reason and understand the question and information better, it also exposed them to making errors due to incorrect/missing domain specific knowledge as well as reasoning and calculation errors. Additionally, we noticed that FS helped LLMs the most in both Levels thanks to the integration of positive instances into the prompt, yielding the best performance in most cases.

With these observations in mind, we propose future systems that could display greater performance by utilizing various tools. The most prevalent error mode of CoT, knowledge errors, could be addressed through retrieval-augmented generation using an external knowledge base containing CFA-specific information. Calculation errors could be avoided by offloading calculations to a function or API such as Wolfram Alpha. The remaining error modes, reasoning and inconsistency, could be reduced by employing a critic model to review and second guess the thinking before submitting the answer, or combining FS and CoT together to give richer examples of expected behavior. We hope this work paves the way for future studies to continue enhancing LLMs for financial reasoning problems through rigorous evaluation.

\section*{Acknowledgments}
This research was funded in part by the Faculty
Research Awards of J.P. Morgan AI Research. The
authors are solely responsible for the contents of the paper and the opinions expressed in this publication do not reflect those of the funding agencies.

\textbf{Disclaimer} This paper was prepared for informational purposes by the Artificial Intelligence Research group of JPMorgan Chase \& Co and its affiliates (``JP Morgan”), and is not a product of the Research Department of JP Morgan. JP Morgan makes no representation and warranty whatsoever and disclaims all liability, for the completeness, accuracy or reliability of the information contained herein. This document is not intended as investment research or investment advice, or a recommendation, offer or solicitation for the purchase or sale of any security, financial instrument, financial product or service, or to be used in any way for evaluating the merits of participating in any transaction, and shall not constitute a solicitation under any jurisdiction or to any person, if such solicitation under such jurisdiction or to such person would be unlawful.

% The acknowledgments should go immediately before the references.  Do
% not number the acknowledgments section. Do not include this section
% when submitting your paper for review.

% % include your own bib file like this:
\bibliographystyle{acl}
\bibliography{acl}

\begin{thebibliography}{}

\bibitem[\protect\citename{Abdaljalil and
  Bouamor}2021]{abdaljalil2021exploration}
Samir Abdaljalil and Houda Bouamor.
\newblock 2021.
\newblock An exploration of automatic text summarization of financial reports.
\newblock In {\em Proceedings of the Third Workshop on Financial Technology and
  Natural Language Processing}, pages 1--7.

\bibitem[\protect\citename{Araci}2019]{araci2019finbert}
Dogu Araci.
\newblock 2019.
\newblock Finbert: Financial sentiment analysis with pre-trained language
  models.
\newblock {\em arXiv preprint arXiv:1908.10063}.

\bibitem[\protect\citename{Bang \bgroup et al.\egroup }2023]{bang2023multitask}
Yejin Bang, Samuel Cahyawijaya, Nayeon Lee, Wenliang Dai, Dan Su, Bryan Wilie,
  Holy Lovenia, Ziwei Ji, Tiezheng Yu, Willy Chung, Quyet~V. Do, Yan Xu, and
  Pascale Fung.
\newblock 2023.
\newblock A multitask, multilingual, multimodal evaluation of chatgpt on
  reasoning, hallucination, and interactivity.

\bibitem[\protect\citename{Barros \bgroup et al.\egroup
  }2023]{barros2023leveraging}
Thierry~S Barros, Carlos Eduardo~S Pires, and Dimas~Cassimiro Nascimento.
\newblock 2023.
\newblock Leveraging bert for extractive text summarization on federal police
  documents.
\newblock {\em Knowledge and Information Systems}, pages 1--31.

\bibitem[\protect\citename{Brown \bgroup et al.\egroup
  }2020]{brown2020language}
Tom Brown, Benjamin Mann, Nick Ryder, Melanie Subbiah, Jared~D Kaplan, Prafulla
  Dhariwal, Arvind Neelakantan, Pranav Shyam, Girish Sastry, Amanda Askell,
  et~al.
\newblock 2020.
\newblock Language models are few-shot learners.
\newblock {\em Advances in neural information processing systems},
  33:1877--1901.

\bibitem[\protect\citename{Chen \bgroup et al.\egroup }2022]{chen2022convfinqa}
Zhiyu Chen, Shiyang Li, Charese Smiley, Zhiqiang Ma, Sameena Shah, and
  William~Yang Wang.
\newblock 2022.
\newblock Convfinqa: Exploring the chain of numerical reasoning in
  conversational finance question answering.

\bibitem[\protect\citename{Chowdhery \bgroup et al.\egroup
  }2022]{chowdhery2022palm}
Aakanksha Chowdhery, Sharan Narang, Jacob Devlin, Maarten Bosma, Gaurav Mishra,
  Adam Roberts, Paul Barham, Hyung~Won Chung, Charles Sutton, Sebastian
  Gehrmann, et~al.
\newblock 2022.
\newblock Palm: Scaling language modeling with pathways.

\bibitem[\protect\citename{Frieder \bgroup et al.\egroup
  }2023]{frieder2023mathematical}
Simon Frieder, Luca Pinchetti, Alexis Chevalier, Ryan-Rhys Griffiths, Tommaso
  Salvatori, Thomas Lukasiewicz, Philipp~Christian Petersen, and Julius Berner.
\newblock 2023.
\newblock Mathematical capabilities of chatgpt.

\bibitem[\protect\citename{Katz \bgroup et al.\egroup }2023]{katz2023gpt}
Daniel~Martin Katz, Michael~James Bommarito, Shang Gao, and Pablo Arredondo.
\newblock 2023.
\newblock Gpt-4 passes the bar exam.
\newblock {\em Available at SSRN 4389233}.

\bibitem[\protect\citename{Khashabi \bgroup et al.\egroup
  }2020]{khashabi2020unifiedqa}
Daniel Khashabi, Sewon Min, Tushar Khot, Ashish Sabharwal, Oyvind Tafjord,
  Peter Clark, and Hannaneh Hajishirzi.
\newblock 2020.
\newblock Unifiedqa: Crossing format boundaries with a single qa system.

\bibitem[\protect\citename{Kim \bgroup et al.\egroup }2022]{kim2022ocr}
Geewook Kim, Teakgyu Hong, Moonbin Yim, JeongYeon Nam, Jinyoung Park, Jinyeong
  Yim, Wonseok Hwang, Sangdoo Yun, Dongyoon Han, and Seunghyun Park.
\newblock 2022.
\newblock Ocr-free document understanding transformer.
\newblock In {\em European Conference on Computer Vision}, pages 498--517.
  Springer.

\bibitem[\protect\citename{Kung \bgroup et al.\egroup
  }2023]{kung2023performance}
TH~Kung, M~Cheatham, A~Medenilla, C~Sillos, L~De~Leon, C~Elepa{\~n}o, et~al.
\newblock 2023.
\newblock Performance of chatgpt on usmle: Potential for ai-assisted medical
  education using large language models. plos digit health 2 (2): e0000198.

\bibitem[\protect\citename{Kıcıman \bgroup et al.\egroup
  }2023]{kiciman2023causal}
Emre Kıcıman, Robert Ness, Amit Sharma, and Chenhao Tan.
\newblock 2023.
\newblock Causal reasoning and large language models: Opening a new frontier
  for causality.

\bibitem[\protect\citename{Le \bgroup et al.\egroup }2022]{le2022coderl}
Hung Le, Yue Wang, Akhilesh~Deepak Gotmare, Silvio Savarese, and Steven
  Chu~Hong Hoi.
\newblock 2022.
\newblock Coderl: Mastering code generation through pretrained models and deep
  reinforcement learning.
\newblock {\em Advances in Neural Information Processing Systems},
  35:21314--21328.

\bibitem[\protect\citename{Lewis \bgroup et al.\egroup
  }2020]{lewis2020retrieval}
Patrick Lewis, Ethan Perez, Aleksandra Piktus, Fabio Petroni, Vladimir
  Karpukhin, Naman Goyal, Heinrich K{\"u}ttler, Mike Lewis, Wen-tau Yih, Tim
  Rockt{\"a}schel, et~al.
\newblock 2020.
\newblock Retrieval-augmented generation for knowledge-intensive nlp tasks.
\newblock {\em Advances in Neural Information Processing Systems},
  33:9459--9474.

\bibitem[\protect\citename{Li \bgroup et al.\egroup }2022]{li2022unified}
Jingye Li, Hao Fei, Jiang Liu, Shengqiong Wu, Meishan Zhang, Chong Teng,
  Donghong Ji, and Fei Li.
\newblock 2022.
\newblock Unified named entity recognition as word-word relation
  classification.

\bibitem[\protect\citename{Li \bgroup et al.\egroup }2023]{li2023chatgpt}
Xianzhi Li, Xiaodan Zhu, Zhiqiang Ma, Xiaomo Liu, and Sameena Shah.
\newblock 2023.
\newblock Are chatgpt and gpt-4 general-purpose solvers for financial text
  analytics? an examination on several typical tasks.
\newblock {\em arXiv preprint arXiv:2305.05862}.

\bibitem[\protect\citename{Mosbach \bgroup et al.\egroup }2023]{mosbach2023few}
Marius Mosbach, Tiago Pimentel, Shauli Ravfogel, Dietrich Klakow, and Yanai
  Elazar.
\newblock 2023.
\newblock Few-shot fine-tuning vs. in-context learning: A fair comparison and
  evaluation.
\newblock {\em arXiv preprint arXiv:2305.16938}.

\bibitem[\protect\citename{Rahimi \bgroup et al.\egroup
  }2023]{rahimi2023exploring}
Saba Rahimi, Tucker Balch, and Manuela Veloso.
\newblock 2023.
\newblock Exploring the effectiveness of gpt models in test-taking: A case
  study of the driver's license knowledge test.
\newblock {\em arXiv preprint arXiv:2308.11827}.

\bibitem[\protect\citename{Strong \bgroup et al.\egroup
  }2023]{strong2023performance}
Eric Strong, Alicia DiGiammarino, Yingjie Weng, Preetha Basaviah, Poonam
  Hosamani, Andre Kumar, Andrew Nevins, John Kugler, Jason Hom, and Jonathan
  Chen.
\newblock 2023.
\newblock Performance of chatgpt on free-response, clinical reasoning exams.
\newblock {\em medRxiv}, pages 2023--03.

\bibitem[\protect\citename{Touvron \bgroup et al.\egroup
  }2023]{touvron2023llama}
Hugo Touvron, Thibaut Lavril, Gautier Izacard, Xavier Martinet, Marie-Anne
  Lachaux, Timoth{\'e}e Lacroix, Baptiste Rozi{\`e}re, Naman Goyal, Eric
  Hambro, Faisal Azhar, et~al.
\newblock 2023.
\newblock Llama: Open and efficient foundation language models.
\newblock {\em arXiv preprint arXiv:2302.13971}.

\bibitem[\protect\citename{Vaswani \bgroup et al.\egroup
  }2017]{vaswani2017attention}
Ashish Vaswani, Noam Shazeer, Niki Parmar, Jakob Uszkoreit, Llion Jones,
  Aidan~N Gomez, {\L}ukasz Kaiser, and Illia Polosukhin.
\newblock 2017.
\newblock Attention is all you need.
\newblock {\em Advances in neural information processing systems}, 30.

\bibitem[\protect\citename{Wang \bgroup et al.\egroup }2022]{wang2022numerical}
Bin Wang, Jiangzhou Ju, Yunlin Mao, Xin-Yu Dai, Shujian Huang, and Jiajun Chen.
\newblock 2022.
\newblock A numerical reasoning question answering system with fine-grained
  retriever and the ensemble of multiple generators for finqa.
\newblock {\em arXiv preprint arXiv:2206.08506}.

\bibitem[\protect\citename{Wang \bgroup et al.\egroup }2023]{wang2023scibench}
Xiaoxuan Wang, Ziniu Hu, Pan Lu, Yanqiao Zhu, Jieyu Zhang, Satyen Subramaniam,
  Arjun~R Loomba, Shichang Zhang, Yizhou Sun, and Wei Wang.
\newblock 2023.
\newblock Scibench: Evaluating college-level scientific problem-solving
  abilities of large language models.
\newblock {\em arXiv preprint arXiv:2307.10635}.

\bibitem[\protect\citename{Wei \bgroup et al.\egroup }2022]{wei2022chain}
Jason Wei, Xuezhi Wang, Dale Schuurmans, Maarten Bosma, Fei Xia, Ed~Chi, Quoc~V
  Le, Denny Zhou, et~al.
\newblock 2022.
\newblock Chain-of-thought prompting elicits reasoning in large language
  models.
\newblock {\em Advances in Neural Information Processing Systems},
  35:24824--24837.

\bibitem[\protect\citename{Wu \bgroup et al.\egroup }2023]{wu2023bloomberggpt}
Shijie Wu, Ozan Irsoy, Steven Lu, Vadim Dabravolski, Mark Dredze, Sebastian
  Gehrmann, Prabhanjan Kambadur, David Rosenberg, and Gideon Mann.
\newblock 2023.
\newblock Bloomberggpt: A large language model for finance.
\newblock {\em arXiv preprint arXiv:2303.17564}.

\bibitem[\protect\citename{Yang \bgroup et al.\egroup
  }2023a]{yang2023exploitgen}
Guang Yang, Yu~Zhou, Xiang Chen, Xiangyu Zhang, Tingting Han, and Taolue Chen.
\newblock 2023a.
\newblock Exploitgen: Template-augmented exploit code generation based on
  codebert.
\newblock {\em Journal of Systems and Software}, 197:111577.

\bibitem[\protect\citename{Yang \bgroup et al.\egroup }2023b]{yang2023fingpt}
Hongyang Yang, Xiao-Yang Liu, and Christina~Dan Wang.
\newblock 2023b.
\newblock Fingpt: Open-source financial large language models.
\newblock {\em arXiv preprint arXiv:2306.06031}.

\bibitem[\protect\citename{Zhang \bgroup et al.\egroup
  }2023]{zhang2023extractive}
Haopeng Zhang, Xiao Liu, and Jiawei Zhang.
\newblock 2023.
\newblock Extractive summarization via chatgpt for faithful summary generation.
\newblock {\em arXiv preprint arXiv:2304.04193}.

\end{thebibliography}

\clearpage

\pagebreak

\section*{Appendix}

\renewcommand{\thesubsection}{\Alph{subsection}}
\setcounter{subsection}{0}

\subsection{Topic Distribution in each Level}

\begin{figure}[h]
    \centering
    \includegraphics[width=\columnwidth]{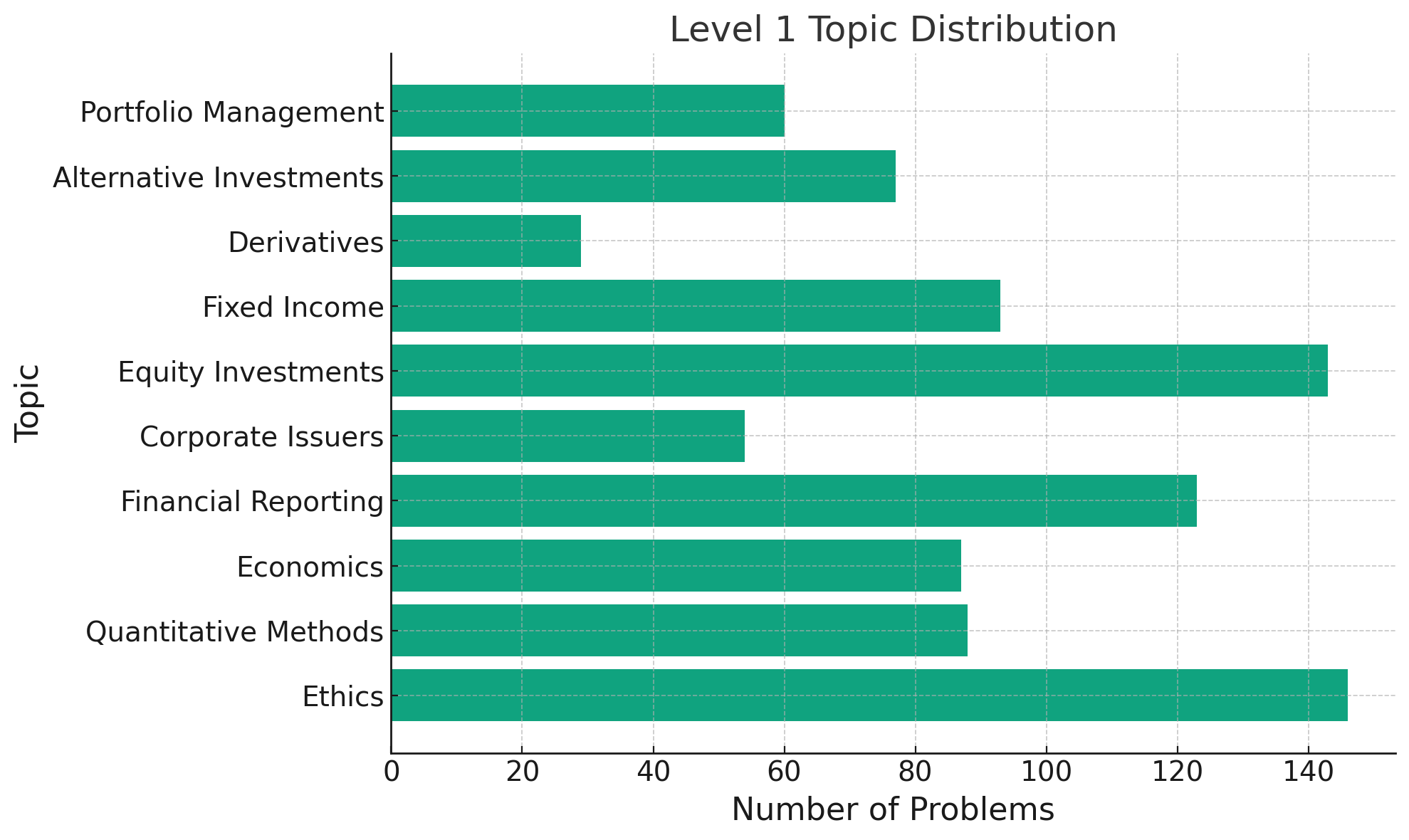}  
    \caption{Level I exam topic distribution}
    \label{fig:l1-topic-dist}
\end{figure}

\begin{figure}[h]
    \centering
    \includegraphics[width=\columnwidth]{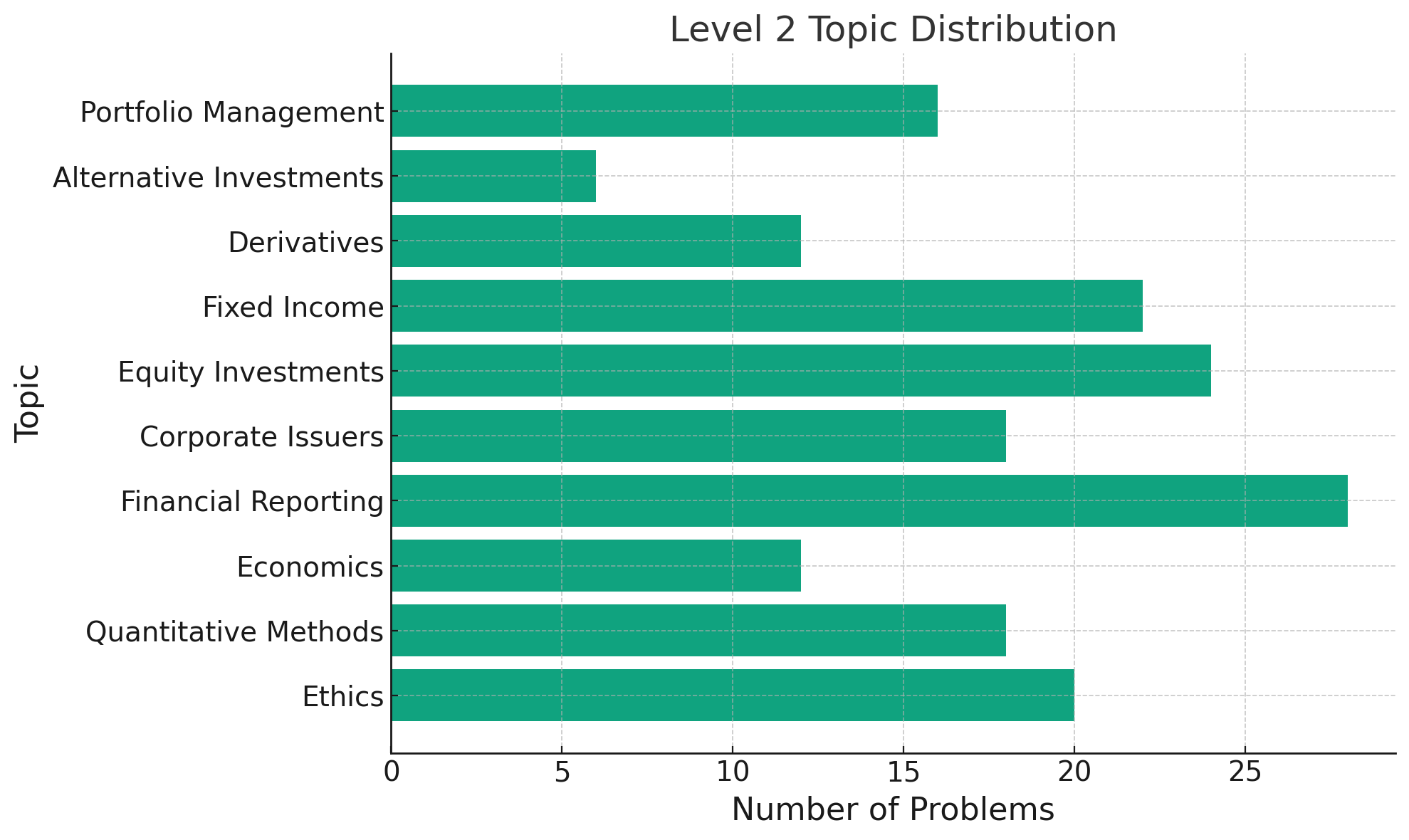}  
    \caption{Level II exam topic distribution}
    \label{fig:l2-topic-dist}
\end{figure}

\subsection{Prompt templates used}

\subsubsection{Level I}

%\paragraph{ZS}
\small
%\begin{minted}[breaklines,breaksymbolleft=]{text}
\begin{lstlisting}[language=TeX, caption={ZS}]
SYSTEM: You are a CFA (chartered financial analyst) taking a test to evaluate your knowledge of finance. You will be given a question along with three possible answers (A, B, and C).

Indicate the correct answer (A, B, or C).

USER: Question:
{question}
A. {choice_a}
B. {choice_b}
C. {choice_c}
\end{lstlisting}
%\end{minted}

%\paragraph{CoT}
%\small
%\begin{minted}[breaklines,breaksymbolleft=]{text}

\begin{lstlisting}[language=TeX, caption={CoT}]
SYSTEM: You are a CFA (chartered financial analyst) taking a test to evaluate your knowledge of finance. You will be given a question along with three possible answers (A, B, and C).

Before answering, you should think through the question step-by-step. Explain your reasoning at each step towards answering the question. If calculation is required, do each step of the calculation as a step in your reasoning.

Indicate the correct answer (A, B, or C).

USER: Question:
{question}
A. {choice_a}
B. {choice_b}
C. {choice_c}
\end{lstlisting}
%\end{minted}

%\paragraph{FS (2S example)}
%\small
%\begin{minted}[breaklines,breaksymbolleft=]{text}
%SYSTEM: You are a CFA (chartered financial analyst) taking a test to evaluate your knowledge of finance.
%You will be given a question along with three possible answers (A, B, and C).

%Indicate the correct answer (A, B, or C).

%USER: Question:
%{question}
%A. {choice_a}
%B. {choice_b}
%C. {choice_c}

%ASSISTANT: {answer}

%USER: Question:
%{question}
%A. {choice_a}
%B. {choice_b}
%C. {choice_c}

%ASSISTANT: {answer}

%USER: Question:
%{question}
%A. {choice_a}
%B. {choice_b}
%C. {choice_c}
%\end{minted}

\begin{lstlisting}[language=TeX, caption={FS (2S example)}]
SYSTEM: You are a CFA (chartered financial analyst) taking a test to evaluate your knowledge of finance. You will be given a question along with three possible answers (A, B, and C).

Indicate the correct answer (A, B, or C).

USER: Question:
{question}
A. {choice_a}
B. {choice_b}
C. {choice_c}

ASSISTANT: {answer}

USER: Question:
{question}
A. {choice_a}
B. {choice_b}
C. {choice_c}

ASSISTANT: {answer}

USER: Question:
{question}
A. {choice_a}
B. {choice_b}
C. {choice_c}
\end{lstlisting}

\subsubsection{Level II}
\normalsize
For Level II, the case description of each item-set was inserted before each question from the user.

% \subsection{Memorization Test Example}
% \label{sect:memtest}

% The system message used for memorization tests:

% \begin{minted}[breaklines,breaksymbolleft=]{text}
% SYSTEM: You are an AI assistant that has read many sources of text from the internet. I am looking at text from CFA mock exams, published by . I am going to list some sample questions, and I want you to complete each question text as best as you can. I am testing your memory.
% \end{minted}

% Question:
% \begin{minted}[breaklines,breaksymbolleft=]{text}
% <p>Portfolio ABC has a beta of 1.6 and generated a return of 21%. If the risk-free rate is 2% and the market premium is 10%, Jensen\u2019s alpha for this portfolio is <em>closest to</em>:</p>
% \end{minted}
 
% Prompt:
% \begin{minted}[breaklines,breaksymbolleft=]{text}
% <p>Portfolio ABC has a beta of 1.6 and generated a return of 21%. If the risk-free rate is
% \end{minted}
 
% Response:
% \begin{minted}[breaklines,breaksymbolleft=]{text}
% 5%, what is the expected return of the market?</p>

% To calculate the expected return of the market, we can use the Capital Asset Pricing Model (CAPM) formula. The CAPM formula is as follows: 
% ...
% \end{minted}

\end{document}